\documentclass[journal]{IEEEtran}

\usepackage[T1]{fontenc}
\usepackage[utf8]{inputenc}
\usepackage{cite}
\usepackage{url}
\usepackage{graphicx}
\usepackage{booktabs}
\usepackage{multirow}

\begin{document}
	
	\title{Hierarchical Federated Learning for Networked AI: From Communication Saving to Architecture-Aware Design}
	
	\author{Seyed~Mohammad~Azimi-Abarghouyi,~\IEEEmembership{Member,~IEEE}, Mehdi~Bennis,~\IEEEmembership{Fellow,~IEEE}, and Leandros Tassiulas,~\IEEEmembership{Fellow,~IEEE}
		\thanks{S. M. Azimi-Abarghouyi is with the Department of Electrical Engineering, Chalmers University of Technology, Gothenburg, Sweden (e-mail: azimimo@chalmers.se).
			
			M. Bennis is with the Centre for Wireless Communications, University of
			Oulu, Oulu, Finland (e-mail: mehdi.bennis@oulu.fi).
			
			L. Tassiulas is with the Department of Electrical Engineering,
			Yale University, New Haven, USA
			(e-mail: leandros.tassiulas@yale.edu). 
			
			}	}
	
	\maketitle
	
	\begin{abstract}
		Federated learning (FL) is fundamentally a distributed optimization problem executed by communicating agents with local data, local computation, and partial system visibility. Once FL is viewed through that lens, hierarchy is not merely a scalability mechanism. It becomes the natural place to rethink how distributed optimization should be organized over real multi-tier networks. This article argues that hierarchical federated learning (HFL) should move beyond its common framing as a communication-saving protocol and instead be viewed as an architecture-aware design framework for networked AI. The framework is organized around three coupled design axes: architectural parameters, layer-wise optimization decomposition, and layer-wise communication realization. The first axis determines the coordination geometry of learning through hierarchy depth, layer asymmetry, and layered connectivity. The second determines how the global FL objective is decomposed across layers and highlights modular multi-layer optimization as a major opportunity beyond one dominant method everywhere. The third determines how the distributed optimization is physically realized under heterogeneous communication regimes, from interference-limited lower tiers to reliable upper tiers. A central message is that, in HFL, convergence becomes architecture-dependent: it is directly shaped by the chosen hierarchy, the assigned optimization roles, and the communication mechanisms that connect them. We develop this viewpoint using large-scale wireless edge intelligence as a flagship networked AI setting, then provide a comparative perspective on flat FL, two-tier HFL, and deep HFL together with a regime-oriented design map. The resulting perspective positions HFL as a practical methodology for designing future networked AI systems.
	\end{abstract}
	
	\begin{IEEEkeywords}
		Federated learning, hierarchical federated learning, networked AI, edge intelligence, distributed optimization, multi-tier networks.
	\end{IEEEkeywords}
	\vspace{-5pt}
	\section{Introduction}
	
	Federated learning (FL) has become a central paradigm for distributed intelligence because it allows many devices to train a shared model without transferring raw data to a central location \cite{McmahanML2017}. This principle is attractive for privacy, regulation, and communication efficiency, and it has motivated interest across wireless networks, Internet-of-Things systems, industrial automation, vehicular services, enterprise platforms, and edge computing. Yet the canonical FL picture remains simple: many devices exchange updates with one coordinating server.
	
	That picture becomes restrictive once learning is embedded into real networked systems. Practical deployments are rarely flat. They already contain several coordination layers, such as gateways, access points, edge servers, regional controllers, data centers, and cloud resources. Once learning is deployed over such infrastructures, performance is shaped not only by the optimizer, but also by the communication architecture that transports, distorts, aggregates, and redistributes information \cite{VirginiaSPM2020}. This is why hierarchical federated learning (HFL) has attracted growing interest \cite{LetaiefTWC2023}. Clustered FL is also related, as it groups clients according to data or task similarity, offering a complementary perspective \cite{SattlerTNNLS2020}.
	
	However, HFL is still often framed too narrowly as a deeper version of FL with one intermediate device--edge--cloud split and reduced communication load as the main benefit. That view is useful, but incomplete. Practical multi-tier systems can have richer coordination structure, broader asymmetry, and more varied communication regimes. Once hierarchy is introduced, the learning system changes where coordination occurs, how errors propagate, where privacy can be enforced, and how optimization behavior can be distributed across scales. The result is not merely a deeper protocol, but an architecture whose behavior emerges from networking, communication, privacy, and optimization.
	
	The main claim of this article is that \emph{HFL should be designed as an architecture-aware framework for networked AI}. FL is not only a learning procedure. It is a distributed optimization problem solved by communicating agents embedded in a network. Architecture therefore shapes what can be coordinated, where information can be fused, and how global intelligence can emerge from local interaction.
	
	Under this perspective, the design question is not simply whether hierarchy should be used, but how distributed optimization should be organized over a general multi-tier architecture. We develop that question through three coupled axes:
	\begin{itemize}
		\item \textbf{Architectural parameters}: the hierarchy depth, layer asymmetry, and layered connection graph that define the coordination geometry of the distributed optimization;
		\item \textbf{Layer-wise optimization decomposition}: how the global FL objective is distributed across layers and which optimization roles or methods are assigned to each layer;
		\item \textbf{Layer-wise communication realization}: how the distributed optimization is physically carried out under different communication regimes across the hierarchy.
	\end{itemize}
	
	This organization leads to the central message of the paper: in HFL, convergence is shaped by architecture-aware design itself. Architectural parameters determine the coordination scales through which information propagates, layer roles determine how optimization is distributed across those scales, and communication realization determines the distortions, delays, and aggregation semantics under which updates are exchanged.
	
	To make this viewpoint concrete, the article anchors the discussion in \emph{large-scale wireless edge intelligence}, a setting that combines dense wireless access, multi-cluster interference, heterogeneous compute resources, partial trust across domains, and the possibility of deeper edge--regional--cloud or space--air--ground hierarchies \cite{AluVTM2021}. In such systems, flat FL is often too rigid, while simplistic two-tier HFL can remain mismatched to the actual deployment.
	
	\begin{figure}[t]
		\centering
		\includegraphics[width=0.383\textwidth]{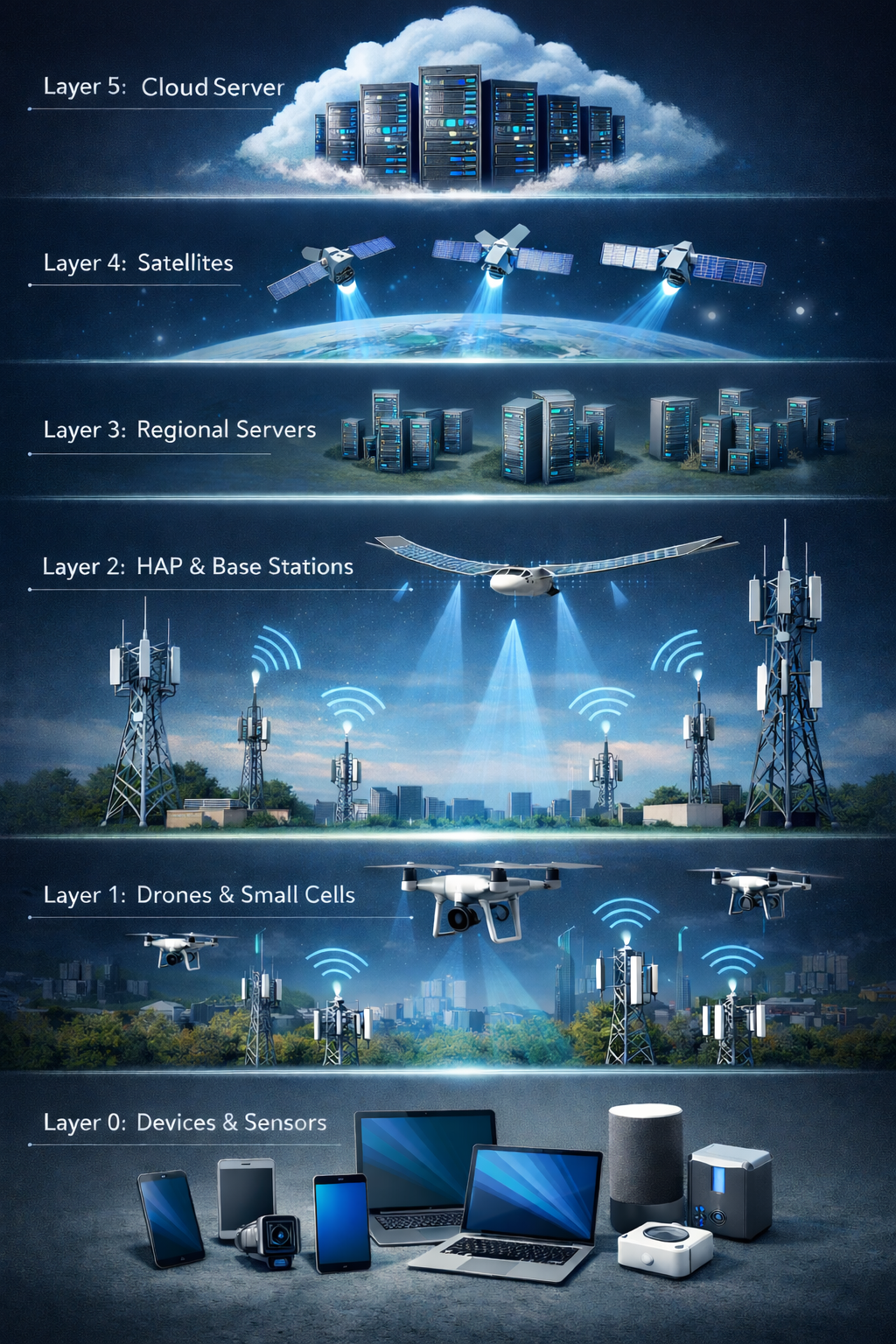}
		\caption{A representative multi-tier network architecture. The system is organized hierarchically across six layers, including cloud servers at Layer~5, satellites at Layer~4, regional servers at Layer~3, macro base stations and a high-altitude platform at Layer~2, small base stations and drones at Layer~1, and end-user devices and sensors at Layer~0. In architecture-aware networked AI, the learning system should follow such structure rather than assume a one-size-fits-all topology.}
		\label{fig:multi_layer_network}
		\vspace{-5pt}
	\end{figure}
	
	\section{Why Hierarchy Changes the Design Problem}
	
	Hierarchy matters because it changes the object being designed. In flat FL, the usual design target is one distributed optimizer deployed over a network. In HFL, the design target becomes a \emph{network-shaped optimizer}, whose behavior depends on multiple coordination scales, unequal roles, and heterogeneous transport regimes. The question is no longer only how to reduce communication cost around a fixed algorithm, but how the algorithm itself should be organized once the network is layered.
	
	This shift can be diagnosed through three mismatches.
	
	The first is a \textbf{topology mismatch}. Flat FL assumes one dominant split: local computation at devices and global aggregation at one server. Real systems rarely look like that, and many HFL formulations still assume only one intermediate layer between devices and the cloud. That simplification may be analytically convenient, but it can remain structurally misaligned with practical systems in which coordination is already distributed across several layers.
	
	The second is an \textbf{optimization mismatch}. Most FL systems still rely on one dominant optimization logic everywhere, usually local gradient descent plus averaging. But once the network becomes multi-tier, different layers see different information, have different capabilities, and operate under different costs. Imposing one uniform logic across all scales wastes the asymmetry that hierarchy creates.
	
	The third is a \textbf{transport mismatch}. Lower wireless tiers, intermediate edge tiers, and higher backhaul tiers do not create the same communication problem. They differ in interference, reliability, bandwidth, synchronization, and trust. Treating them with one communication view leads to avoidable distortion and poor burden placement.
	
	These mismatches explain why HFL should not be understood merely as FL with extra aggregators. It should be understood as a way of organizing distributed optimization over a general network architecture. The architectural axis addresses the topology mismatch, the optimization axis addresses the algorithmic mismatch, and the communication axis addresses the transport mismatch.
	
	\section{Axis I: Architectural Parameters}
	
	The first axis concerns the architecture itself. In HFL, the network is not a passive substrate on which learning is later implemented. It defines the coordination geometry of the distributed optimization. Three architectural parameters are especially important: \emph{hierarchy depth}, \emph{layer asymmetry}, and the \emph{layered connection graph}. The first determines how many coordination scales exist, the second determines how different those scales are, and the third determines how coordination can actually take place across and within them.
	\vspace{-5pt}
	\subsection{Hierarchy Depth}
	
	Hierarchy depth determines how many stages of coordination exist between end devices and the most global coordination layer. In practice, the right depth is not an abstract preference for shallow or deep learning. It is a question of topology fit.
	
	In large-scale wireless edge intelligence, a shallow hierarchy is attractive because it keeps the system simple. Fewer layers mean fewer interfaces, fewer synchronization points, and fewer opportunities for delay and distortion to accumulate. When devices are naturally grouped around capable edge servers and the edge--cloud path is reasonably strong, a two-tier design may already capture much of the benefit of HFL.
	
	However, shallow simplicity becomes misleading when the network itself is not shallow. Devices may connect to local access points, those access points to edge servers, and those servers to regional platforms before reaching the cloud. More complex infrastructures can also involve aerial or non-terrestrial components with distinct latency and reliability characteristics \cite{AluVTM2021}. If the learning architecture compresses all of this into one intermediate layer, the result can be overloaded coordinators, inefficient long-range interaction, or fragile cross-domain synchronization.
	
	This is why two-tier HFL should not be treated as a universal default. In the standard two-tier view, the architecture is represented through one intermediate layer between devices and the cloud. That abstraction is appropriate only when one such layer matches the dominant coordination split of the deployment. Otherwise, the learning architecture becomes simpler than the system it is meant to organize.
	
	Recent generalized models of HFL reinforce this point by showing that the number of layers is itself a meaningful architectural variable in convergence behavior \cite{AzimiTON2026}. This leads to a useful design insight: hierarchy depth should be chosen by identifying how many \emph{distinct coordination scales} the deployment actually contains. In that sense, depth is not a cosmetic property of HFL. It is the variable that decides the vertical structure of learning over the network.
	
	A complementary insight is that too little depth and too much depth can both be harmful. Too little depth causes \emph{architectural underfitting}: the learning system ignores genuine coordination scales already present in the network. Too much depth causes \emph{architectural overfitting}: the system inserts coordination stages whose cost and complexity are not justified by the deployment. Good HFL design therefore requires matching the learning hierarchy to the actual control hierarchy of the system.
	\vspace{-5pt}
	\subsection{Layer Asymmetry}
	
	Depth alone is not enough. Real networks are asymmetric. Layers differ in computation, visibility, trust, and communication cost. A lower wireless tier may only be able to execute simple local steps and exchange noisy partial states. An intermediate edge tier may have enough capability to stabilize local drift or reconcile several client groups. A higher tier may operate under stronger reliability and broader visibility, and can therefore enforce wider consistency or longer-horizon corrections.
	
	This asymmetry is not a minor implementation detail. It changes what each layer can realistically contribute to solving the FL problem. A layer with limited computation and weak links should not be assigned the same role as a layer with broader visibility, stronger transport, or greater trust. Treating all layers as if they were algorithmically interchangeable wastes the very structure that hierarchy creates.
	
	A useful way to view this is that architecture acts as an \emph{optimization prior}. It constrains which roles are natural, which are expensive, and which are even meaningful at each scale. The same global FL objective can therefore induce different effective algorithms depending on whether the architecture is balanced or skewed, computation-rich or bandwidth-starved, trust-uniform or trust-segmented. In this sense, architectural asymmetry is not a nuisance around the learning problem. It is part of the learning problem itself.
	\vspace{-5pt}
	\subsection{Layered Connection Graphs}
	
	The third parameter is the layered connection graph. Even when the number of layers is fixed and their asymmetry is understood, the coordination structure is still not fully specified unless one knows how nodes are connected across and within layers.
	
	This graph is rarely regular. Some intermediate nodes connect to many lower-tier nodes, others to few. Some domains are strongly coupled, others weakly linked. In addition, useful topologies are not necessarily purely vertical trees. There may be horizontal edges within a layer, allowing peer nodes to exchange information before communicating across layers.
	
	This matters because the graph determines which coordination patterns are feasible. A tree-like structure naturally supports upward fusion and downward dissemination. A layer with lateral edges can additionally support peer-to-peer refinement, local consensus, or neighborhood mixing before information moves to a broader scale. Architecture here therefore does not imply that every layer must coordinate only through a single parent node. It means that coordination is organized across ordered layers or scales, while the precise interactions within a scale are determined by the graph that exists there.
	
	The layered connection graph also determines which coordination forms are structurally plausible at a given scale. Depending on the available links, a layer may support centralized coordination through designated parent nodes, decentralized coordination among neighboring nodes, or hybrid forms that combine intra-layer mixing with inter-layer aggregation. In this sense, architecture does not merely say who is connected to whom. It determines which coordination patterns the distributed optimizer can actually realize.
	
	This point is also important analytically. Once HFL is generalized beyond the standard two-tier abstraction, convergence need not depend only on how many layers exist; it also depends on how nodes are connected across and within those layers \cite{AzimiTON2026}. In that richer view, the layered graph is not merely an implementation detail. It becomes part of the architectural characterization of learning itself. By contrast, the standard two-tier view represents architecture much more coarsely and therefore does not explicitly capture such detailed graph dependence.
	
	Taken together, depth, asymmetry, and layered connectivity define the architecture-dependent search space in which HFL operates. They do not yet say \emph{how} the FL objective should be solved, but they determine the structural possibilities within which that solution must be designed.
	
	\section{Axis II: Layer-Wise Optimization Decomposition}
	
	If the first axis decides the coordination geometry, the second decides how the global FL objective is actually distributed across that geometry. This is where one of the deepest opportunities of HFL emerges.
	
	In most FL systems, the entire network is governed by one dominant optimization logic, typically local gradient descent at devices followed by averaging at higher levels. Even when the architecture becomes hierarchical, the same optimization behavior is often replicated across all layers \cite{LetaiefTWC2023}. This uniformity is convenient, but it also hides one of the most important consequences of hierarchy: different layers do not have the same capabilities, do not see the same information, and do not operate under the same communication constraints. There is no strong reason they should all solve the global FL problem in the same way.
	
	HFL therefore opens a new way to think about distributed optimization in FL itself. Instead of relying on one method everywhere, one can assign different optimization roles to different layers, and potentially use different optimization methods across those layers. In this view, hierarchy is not only a communication scaffold. It is a \emph{multi-layer optimization architecture}.
	\vspace{-5pt}
	\subsection{Why Layer-Wise Optimization Matters}
	
	The first nontrivial consequence of hierarchy is that optimization budgets become multi-scale. In flat FL, one usually chooses local step counts, participation rates, and global aggregation frequencies. In HFL, those budgets are distributed across layers. A lower tier may execute many fast but noisy updates. An intermediate tier may aggregate or correct them at a slower but more informed timescale. A higher tier may update even less frequently while influencing the system more broadly. Hierarchy therefore changes not only where aggregation occurs, but also what counts as a meaningful ``step'' of the global optimizer.
	
	The second consequence is that hierarchy changes the \emph{semantics} of the exchanged object. One layer may naturally operate on gradients, another on model parameters, another on residual-like corrections or structured consensus variables. These are not merely different message formats. They imply different optimization behaviors, different robustness properties, and different sensitivities to communication distortion.
	
	The third consequence is that hierarchy creates room for algorithmic complementarity. A method that is ideal at one scale may be inappropriate at another. For example, a lower layer may need very light computation, while an upper layer can afford more structured coordination. This complementarity is fundamentally unavailable in flat FL, where the architecture provides only one dominant scale of coordination.
	\vspace{-5pt}
	\subsection{A Dictionary of FL Algorithms Across Layers}
	
	Different optimization methods exhibit distinct characteristics:
	\begin{itemize}
		\item \textbf{Gradient descent} provides algorithmic simplicity and low computational overhead.
		\item \textbf{Variance-reduced methods} enhance statistical efficiency by mitigating gradient noise.
		\item \textbf{Proximal methods} improve stability in the presence of regularization and constraints.
		\item \textbf{ADMM} facilitates structured consensus and allows explicit control over agreement across distributed variables.
		\item \textbf{Second-order methods} accelerate convergence when curvature information can be effectively exploited, albeit at the cost of additional computational complexity.
	\end{itemize}
	
	Once a hierarchy exists, these properties do not need to be chosen globally once and for all. They can be distributed across layers and made to complement one another. This is the deeper opportunity of layer-wise optimization: the hierarchy allows algorithmic properties to be \emph{composed} rather than merely repeated.
	
	Our earlier work provides an initial example of this direction \cite{AzimiNL2025}. There, ADMM was used at the top layer and combined with either gradient descent or ADMM at lower layers, leading to two hierarchical FL variants with distinct behaviors. The broader importance of that result is not limited to those specific schemes. Its deeper significance is conceptual: different layers can employ different optimization methods to solve the same FL objective, and their interaction becomes part of the algorithm design itself.
	
	That observation suggests a much broader perspective. In a network with multiple layers, one can in principle assign different optimization methods across the hierarchy, creating many possible algorithmic compositions. Different assignments may yield fundamentally different FL algorithms, even when they target the same global objective. Their behavior depends not only on the methods themselves, but also on how the layers interact through aggregation, synchronization, and information exchange. In this sense, one may even envision a kind of \emph{dictionary} of FL algorithms, structured by how optimization methods are mapped onto layers of the network.
	
	This design space is largely absent in flat FL, where the architecture offers little room for structured composition. HFL, by contrast, turns the hierarchy into an optimization scaffold. Complementary algorithmic properties can be accumulated across layers rather than imposed uniformly throughout the system. Lightweight local progress can be combined with stronger intermediate stabilization and more structured global coordination. In this way, hierarchy changes the question from ``how do we scale one optimizer?'' to ``how do we architect distributed optimization itself?''
	
	\subsection{From Aggregation Layers to Algorithmic Layers}
	
	A useful way to understand this is through a kind of \emph{zooming view} of the hierarchy. From a coarse perspective, one sees only a few major coordination layers. But when one zooms into any one of those layers, that layer can itself be understood as a smaller distributed optimization system with its own optimization state, local coordination rules, exchanged objects, and internal timescale. In that sense, HFL is not merely a vertical chain of aggregators. It is a structured lattice of optimization components distributed across scales. What appears as one block at a higher level may itself contain meaningful optimization structure when viewed more closely.
	
	This layered view also changes the interpretation of what an intermediate layer is. In the flagship wireless-edge setting, an edge layer is not simply a place where local updates are averaged. It can become the place where cluster-level heterogeneity is absorbed, local drift is stabilized, one optimization representation is translated into another, or neighborhood coordination is carried out before information moves across larger scales. A higher layer can impose broader consensus, regularization, or trust-aware coordination. Once these roles are acknowledged, the hierarchy becomes a vehicle for designing distributed optimization rather than merely transporting updates.
	
	Even within one optimizer family, this can create nontrivial and useful choices. A system may aggregate gradients at one layer and models at another, not accidentally, but because those choices behave differently under communication distortions and heterogeneity \cite{AzimiTWC2024,AzimiIOT2026}. More generally, HFL creates a new class of design questions: which optimization behavior belongs at which layer, and which layer-wise assignment produces the right global behavior for a given network condition?
	\vspace{-5pt}
	\subsection{Coordination Form as Part of Optimization Design}
	
	Layer-wise decomposition is not only about \emph{which optimizer} is used. It is also about \emph{how} a layer realizes its optimization role. In particular, a layer may coordinate in three broad forms:
	\begin{itemize}
		\item \textbf{Centralized coordination}: lower-layer nodes send information to designated coordinators or parent nodes, which fuse updates and redistribute them;
		\item \textbf{Decentralized coordination}: peer nodes at the same layer mix information directly with neighbors through local consensus, gossip, or neighborhood exchange;
		\item \textbf{Hybrid coordination}: a layer combines intra-layer peer refinement with inter-layer aggregation.
	\end{itemize}
	
	This point belongs to the core of the optimization axis because the same architectural graph can support several feasible coordination forms, but the choice among them changes the algorithm itself. A layer that performs peer-level refinement before communicating upward does not merely reduce traffic. It changes the optimization state that the next scale receives. Likewise, a layer that relies only on vertical fusion imposes a different bias-variance and synchronization trade-off than one that mixes horizontally first. Coordination form is therefore part of optimization design, not merely an implementation detail.
	
	A useful distinction now becomes visible. The architectural axis determines whether centralized, decentralized, or hybrid coordination is feasible through the available graph structure. The optimization axis determines which of these forms should actually be used to solve the FL problem at a given layer. This separation is important because it clarifies that graph structure enables possibilities, while optimization design selects among them.
	
	\section{Axis III: Layer-Wise Communication Realization}
	
	If the second axis determines \emph{what} optimization is performed at each layer, the third determines \emph{how that optimization is physically realized}. This is where communication enters the solver itself.
	
	In HFL, communication is not only a carrier of optimization states. It changes what those states effectively are. Each layer sees only the information that the communication system allows it to see. This means that communication realization directly shapes the distributed optimization.
	
	In the flagship setting of large-scale wireless edge intelligence, the lower tiers are usually the most communication sensitive. Many devices or edge nodes may need to transmit simultaneously over dense wireless access. Orthogonal access can then create large bandwidth and latency overhead, which is why analog over-the-air aggregation and related simultaneous transmission strategies have become attractive in FL and HFL \cite{DingTWC2020, AzimiTWC2024, AygunTWC2024}. In a single cluster, the idea is elegant: wireless superposition directly forms aggregated information. But the situation becomes more complicated in multi-cluster HFL, where neighboring clusters may operate concurrently and inter-cluster interference becomes a first-order factor \cite{AzimiTWC2024,HuangTWC2021,AygunTWC2024}. A communication method that is appealing within one cluster may become unstable at network scale if multi-cluster coupling is ignored.
	
	The downlink introduces a different challenge. In flat FL, one often assumes that a central server broadcasts a global model to all devices. In HFL, several servers may need to disseminate updates simultaneously to different clusters, potentially over shared resources. Devices may then receive slightly different distorted versions of the intended model. Such inconsistency is not merely an implementation inconvenience. It can alter the effective learning dynamics across clusters and affect convergence.
	
	Upper tiers pose yet another communication regime. As the system moves from dense wireless access to regional or cloud coordination, communication may become less interference limited and more constrained by transport reliability, compression cost, or synchronization overhead. At those layers, digital communication with quantization or scheduled exchange may be more natural than analog aggregation. But upper tiers are not automatically easier. The interface between adjacent layers becomes critical: the representation produced by one layer must be suitable for the communication and processing assumptions of the next.
	
	The design lesson is therefore not to seek one communication mechanism for the whole hierarchy. It is to \emph{match} communication modes to the role and constraints of each layer. Lower wireless tiers may benefit from simultaneous aggregation, interference-aware reception, or local scheduling. Horizontal local interaction may rely on short-range neighborhood exchange. Upper tiers may rely on digital transport, compression, and stronger synchronization. What matters is compatibility across the stack. If architecture and optimization roles are well chosen but communication is mismatched, the distributed optimizer itself is mismatched.
	
	This also connects directly to privacy and trust. In flat FL, privacy is often treated as either device-local perturbation or server-side policy. HFL creates intermediate points where privacy-preserving aggregation and trust-aware communication policies can be placed more efficiently \cite{BrintonTON2026}. This matters because privacy, like communication, does not only constrain the system externally. It changes the effective information available to the optimizer.
	
	\section{Design Synthesis: Architecture-Dependent Convergence}
	
	The previous three sections imply the central message of the paper: in HFL, convergence is architecture-dependent.
	
	In flat FL, convergence discussions often focus on optimizer choice, local update budget, participation level, or communication frequency. Those factors remain important in HFL, but they are no longer sufficient. Once distributed optimization is organized across a hierarchy, convergence is also shaped by the chosen architecture, the optimization roles assigned to layers, and the communication mechanisms that connect them.
	
	Depth determines how many coordination scales updates traverse before becoming global. Layer asymmetry and connection structure determine how information propagates and where bottlenecks emerge. Layer-wise optimization decomposition determines how much progress or stabilization is injected at each scale. Communication realization determines what distortions, delays, and aggregation semantics accompany that progress. Convergence is therefore not only algorithm-dependent; it is \emph{system-dependent} in a much stronger sense.
	
	A useful way to interpret good HFL design is as a problem of \emph{burden placement}. Noise-sensitive aggregation should occur where signals can be fused cheaply and frequently. Stability-enhancing or consensus-enforcing operations should occur where visibility is broader and coordination is less fragile. Privacy operations should be placed where trust and communication budgets make them least destructive. The purpose of hierarchy is therefore not only to split work, but to place each kind of work at the scale where it is most structurally appropriate.
	
	Poor HFL designs tend to fail in recognizable ways. Some place broad coordination too low in the hierarchy, where visibility is limited and communication is fragile. Others place local correction too high, forcing coarse global layers to absorb problems that should have been resolved earlier. Still others choose communication mechanisms that are locally attractive but globally inconsistent with the optimizer they are supposed to support. The value of the framework is that it makes such failures diagnosable before they appear only as disappointing learning curves.
	
	The same viewpoint provides a compact way to compare the main architectural forms of FL. Table~\ref{tab:compare} summarizes the differences among flat FL, two-tier HFL, and deep HFL across dimensions that matter directly in networked AI systems.

	Deep HFL is not automatically the best choice; it helps only when the deployment actually contains several meaningful coordination scales. Two-tier HFL should not be treated as a universal default; it works well only when one intermediate layer captures the dominant topology and burden split. The deeper advantage of HFL is not simply more aggregation stages, but the ability to make the distributed optimization architecture follow the communication architecture.
	
	Table~\ref{tab:matrix} then maps representative operating regimes to suitable HFL choices.

	The regime map highlights four practical lessons. When local wireless access is the dominant burden, communication realization becomes the first design priority. When edge infrastructure is strong, optimization decomposition matters more than added depth. When the network spans several scopes of coordination, architectural parameters become decisive. When the environment is dynamic, the hierarchy itself may need to adapt.

	\begin{table*}[t!]
		\caption{Compact comparative view of flat FL, two-tier HFL, and deep HFL}
		\label{tab:compare}
		\centering
		\footnotesize
		\begin{tabular}{p{2.5cm} p{3.0cm} p{3.1cm} p{3.6cm}}
			\toprule
			\textbf{Dimension} & \textbf{Flat FL} & \textbf{Two-Tier HFL} & \textbf{Deep HFL} \\
			\midrule
			Architecture view & One coordination scale & One dominant local-global split & Several coordination scales with explicit architectural structure \\
			\midrule
			Optimization decomposition & One dominant global coordination logic & Partial separation between local and global roles & High flexibility for assigning distinct optimization methods and semantics across layers \\
			\midrule
			Coordination form & Mostly centralized at one server & Mainly vertical coordination with limited intermediate refinement & Centralized, decentralized, and hybrid forms can coexist across scales \\
			\midrule
			Communication realization & Concentrated at one central server & Split across device--edge and edge--cloud segments & Distributed across multiple tiers with heterogeneous communication regimes \\
			\midrule
			Convergence characterization & Mainly algorithm and participation dependent & Algorithm plus one intermediate coordination scale & Directly architecture dependent: depth, connectivity graph, layer roles, coordination form, and communication jointly matter \\
			\midrule
			Main systems bottleneck & Central traffic concentration and synchronization & Edge overload or rigid edge--cloud coupling & Architecture mismatch, cross-layer interface design, and multi-stage error propagation \\
			\bottomrule
		\end{tabular}
	\end{table*}
	
	\begin{table*}[t!]
		\caption{Illustrative regime-oriented design matrix for networked AI systems}
		\label{tab:matrix}
		\centering
		\footnotesize
		\begin{tabular}{p{3.0cm} p{2.3cm} p{2.4cm} p{3.3cm} p{3.0cm}}
			\toprule
			\textbf{Representative regime} & \textbf{Architectural choice} & \textbf{Coordination form} & \textbf{Optimization emphasis} & \textbf{Communication emphasis} \\
			\midrule
			Dense local wireless clusters with limited spectrum & Shallow hierarchy; two-tier by default, three-tier if an additional regional coordination scale is present & Mostly centralized or hybrid near the edge & Lightweight device updates; edge absorbs local heterogeneity and performs fast local fusion & Interference-aware simultaneous aggregation or carefully scheduled local uplink; compressed digital forwarding upward \\
			\midrule
			Moderate-scale deployment with strong edge servers and weaker cloud backhaul & Two-tier with a strong edge role & Hybrid & Edge carries much of the intermediate coordination and heterogeneity reduction; cloud provides slower global consistency & Rich local coordination plus infrequent but reliable edge--cloud exchange \\
			\midrule
			Wide-area multi-domain system with regional controllers or non-terrestrial relays & Deep hierarchy with explicit regional coordination layers & Centralized across scales, optionally hybrid within regional layers & Optimization roles split by geography and administrative scope; regional layers reconcile cross-domain variation before cloud fusion & Mixed-mode communication: wireless access below, digital transport above \\
			\midrule
			Privacy-sensitive deployment with trusted intermediate infrastructure & Depth follows trust structure rather than traffic alone & Centralized or hybrid & Keep lower-layer learning lightweight; use trusted intermediate layers for protected aggregation and broader consistency & Per-layer privacy and communication policies rather than uniform device-side perturbation \\
			\midrule
			Highly dynamic environment with mobility or varying connectivity & Adaptive or reconfigurable hierarchy & Decentralized or hybrid & Simple device-side behavior with dynamically reassigned intermediate roles & Opportunistic local exchange and semi-synchronous upper-tier coordination \\
			\bottomrule
		\end{tabular}
	\end{table*}

	\section{Implications and Open Directions}
	
	The framework changes how HFL should be approached in both practice and research.
	
	In practice, design should begin from the network itself: what coordination scales exist, how asymmetric the layers are, what communication regimes dominate them, and which layers can support richer optimization roles. Intermediate layers should no longer be justified only as traffic reducers. They may also absorb heterogeneity, create privacy-preserving aggregation points, bridge incompatible communication modes, or support different optimization behaviors. Likewise, HFL changes how bottlenecks are diagnosed: it does not remove them, but redistributes them across layers and interfaces.
	
	For research, several directions stand out.
	
	\textbf{Architecture selection} remains open: most HFL works still assume a fixed hierarchy, whereas future systems need mechanisms that can infer or adapt depth, layer asymmetry, and coordination graphs from system state.
	
	\textbf{Modular optimization theory} is still immature: which optimizer combinations are complementary, and how should one characterize their interaction with hierarchy depth?
	
	\textbf{Coordination-form design} remains underexplored: when should a layer rely on centralized fusion, when should it prefer decentralized peer refinement, and when should it combine the two?
	
	\textbf{Cross-layer control} remains challenging: once layers use different roles and communication modes, the system must decide synchronization rates, compression budgets, privacy placement, and local-global timescale interaction.
	
	\textbf{Evaluation methodology} also deserves attention. If HFL is truly architecture-aware, then comparing schemes only through end-to-end accuracy curves can be misleading. Future evaluation should expose which burden is being relieved, where distortion is being absorbed, and which coordination scale is actually responsible for the gain.
	
	Finally, \textbf{network-aware theory} requires stronger integration among distributed optimization, interference modeling, quantization analysis, privacy composition, and architectural flexibility. The challenge is not to analyze these effects separately, but to understand how they interact once the learning system is distributed across several coordination scales.
	
	\section{Conclusions}
	
Hierarchical federated learning should not be adopted merely as a generic deeper version of FL. It should be designed as a networked AI architecture.
	
	That design problem is fundamentally architecture aware. It requires choosing the right architectural parameters, assigning meaningful optimization roles across layers, and matching communication mechanisms to those roles. Among these axes, modular multi-layer optimization deserves special emphasis because it opens a new way to think about distributed optimization in FL itself. Instead of relying on one dominant method across all layers, HFL makes it possible to distribute and compose optimization behaviors across the network.
	
	The strongest conceptual payoff is therefore not that HFL adds layers to FL. It is that HFL transforms FL from one distributed optimizer with communication constraints into a structured \emph{family} of distributed optimizers shaped by architecture. The key question is no longer ``how many layers should be added?'' but ``which optimization and communication responsibilities belong at which coordination scale?''
	
	When these choices are aligned with the deployment, HFL becomes more than a communication-saving mechanism. It becomes a practical methodology for organizing distributed optimization over multi-tier infrastructures. For practitioners, there is no universally correct hierarchy; the best design is the one that matches the coordination scales, asymmetry, trust structure, and communication constraints of the target system. For researchers, the strongest advances will come from understanding how architecture, communication, and optimization jointly shape the solvability of FL itself.
	
	That is why HFL matters for networked AI. It provides a principled way to turn network structure from a constraint into a source of optimization leverage.

\end{document}